\documentclass{article}
\pdfoutput=1

\PassOptionsToPackage{numbers, compress}{natbib}

\usepackage[preprint]{neurips_2024}

\usepackage[utf8]{inputenc} 
\usepackage[T1]{fontenc}    
\usepackage{url}            
\usepackage{booktabs}       
\usepackage{amsfonts}       
\usepackage{nicefrac}       
\usepackage{microtype}      

\usepackage{subfloat}
\usepackage{graphicx}
\usepackage{booktabs}
\usepackage{multirow}
\usepackage{makecell}
\usepackage{pifont}
\usepackage{wrapfig}
\usepackage{amsmath}
\usepackage{subcaption}
\usepackage{enumitem}
\usepackage[table]{xcolor}
\newcommand{\xmark}{\ding{55}}%

\newcommand*{\img}[1]{%
    \raisebox{-.3\baselineskip}{%
        \includegraphics[
        height=\baselineskip,
        width=\baselineskip,
        keepaspectratio,
        ]{#1}%
    }%
}
\definecolor{citecolor}{HTML}{0071bc}
\usepackage[pagebackref=false,breaklinks=true,letterpaper=true,colorlinks,citecolor=citecolor,bookmarks=false]{hyperref}

\newcommand{\mymodel}{\text{\textsc{VideoLLM-MoD}}}

\definecolor{verylightblue}{RGB}{220,235,245}
\newcommand{\baseline}[1]{\cellcolor{verylightblue}{#1}}

\title{VideoLLM-MoD: Efficient Video-Language Streaming with Mixture-of-Depths Vision Computation}

\author{
\textbf{\normalsize{Shiwei Wu$^{1*}$, Joya Chen$^{2}$\thanks{Equal Contribution}, Kevin Qinghong Lin$^{2}$, Qimeng Wang$^{3}$, Yan Gao$^{3}$, Qianli Xu$^{4}$}}\\
\textbf{\normalsize{Tong Xu$^{1}$, Yao Hu$^{3}$, Enhong Chen$^{1}$, Mike Zheng Shou$^{2}$}}  \vspace{1mm}\\ 
$^{1}$University of Science and Technology of China~~~~~~$^2$Show Lab, National University of Singapore~~~~~~\\
$^3$Xiaohongshu.Inc~~~~~~$^4$Institute for Infocomm Research, A*STAR \\
}
\vspace{-25mm}

\begin{document}

\maketitle
\vspace{-5mm}
\begin{abstract}
A well-known dilemma in large vision-language models (\textit{e.g.}, GPT-4, LLaVA) is that while increasing the number of vision tokens generally enhances visual understanding, it also significantly raises memory and computational costs, especially in long-term, dense video frame streaming scenarios. Although learnable approaches like Q-Former and Perceiver Resampler have been developed to reduce the vision token burden, they overlook the context causally modeled by LLMs (\textit{i.e.}, key-value cache), potentially leading to missed visual cues when addressing user queries. In this paper, we introduce a novel approach to reduce vision compute by leveraging redundant vision tokens ``skipping layers'' rather than decreasing the number of vision tokens. Our method, $\mymodel$, is inspired by mixture-of-depths LLMs and addresses the challenge of numerous vision tokens in long-term or streaming video. Specifically, for each transformer layer, we learn to skip the computation for a high proportion (\textit{e.g.}, 80\%) of vision tokens, passing them directly to the next layer. This approach significantly enhances model efficiency, achieving approximately \textasciitilde42\% time and \textasciitilde30\% memory savings for the entire training. Moreover, our method reduces the computation in the context and avoid decreasing the vision tokens, thus preserving or even improving performance compared to the vanilla model. We conduct extensive experiments to demonstrate the effectiveness of $\mymodel$, showing its state-of-the-art results on multiple benchmarks, including narration, forecasting, and summarization tasks in COIN, Ego4D, and Ego-Exo4D datasets.

\end{abstract}
\vspace{-5mm}
\section{Introduction}
Recent advancements in large language models~\cite{gpt1,gpt2,gpt3,instructgpt,llama1,llama2,llama3,gemini}, particularly with GPT-4o~\cite{gpt4o}, have led many to believe that the development of a J.A.R.V.I.S.-like AI assistant is becoming increasingly feasible. Such an assistant would operate in a streaming manner, remain always-on, and be multimodal to facilitate interaction with users. 

While existing video-based Large Multi-modal Models (LMMs)~\cite{video_chatgpt,videochat,vid2seq,videollama,moviechat,otter,embodiedgpt,antgpt, pllava} have shown significant capabilities in general visual content understanding and reasoning, these models primarily operate in an offline setting, provide response for a few sampled frames within a video in the event-level, which falls short in online settings where there is a need for prompt, concise, and frame-aligned answers for the continuous video frames. The most closet model concept to GPT-4o should be VideoLLM-online~\cite{live}, which continuously receives video frames and provide temporally aligned responses when user queries. 
For instance, in response to a query such as ``remind me when I should add salt", the online assistant ought to evaluate each incoming frame and give temporal-aligned suggestions, taking into account historical visual and linguistic context, rather than merely summarizing the video at the event level.
Consequently, online assistants face significant computational demands as they are required to engage in causal modeling for every frame of the long video, and current approaches~\cite{live} only rely exclusively on the CLS token for each frame, limiting the vision capability to spatial understanding, which is inadequate for scenarios that require fine-grained scene understanding.

\begin{figure}
    \centering
    \includegraphics[width=14cm]{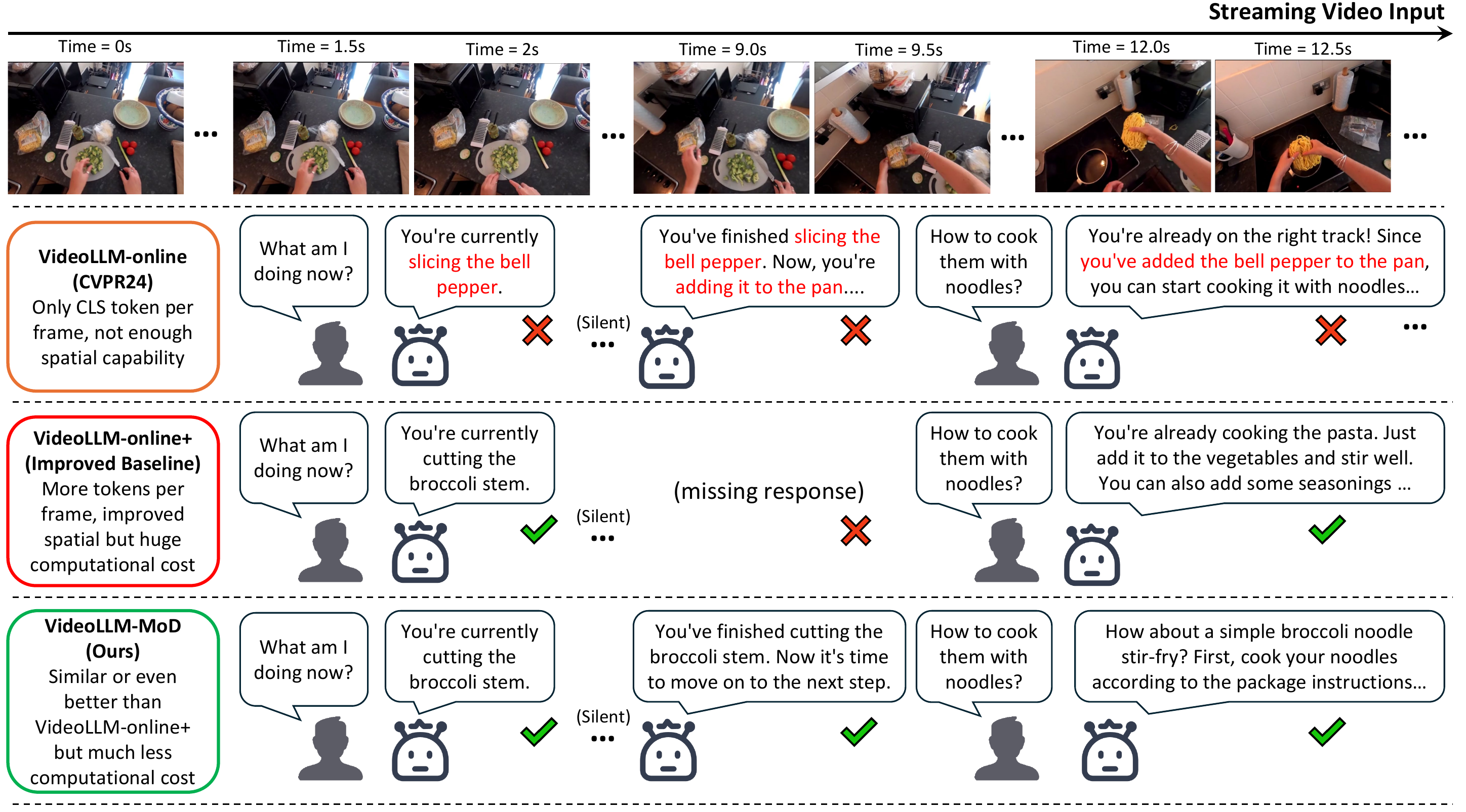}
    \caption{Cases of $\mymodel$ on Ego4D GoalStep~\cite{ego4d_goalstep} video. Using only CLS token often results in spatial understanding errors, \textit{e.g.,} mistaking `broccoli' for `bell pepper.' $\mymodel$ improves fine-grained spatial ability by integrating more spatial tokens while reducing computation costs compared to the improved baseline. Text in \textcolor{red}{red} indicates incorrect response.}
    \label{fig:teaser}
    \vspace{-5mm}
\end{figure}

\begin{wrapfigure}{r}{0.4\textwidth}
    \centering
    \includegraphics[width=0.38\textwidth]{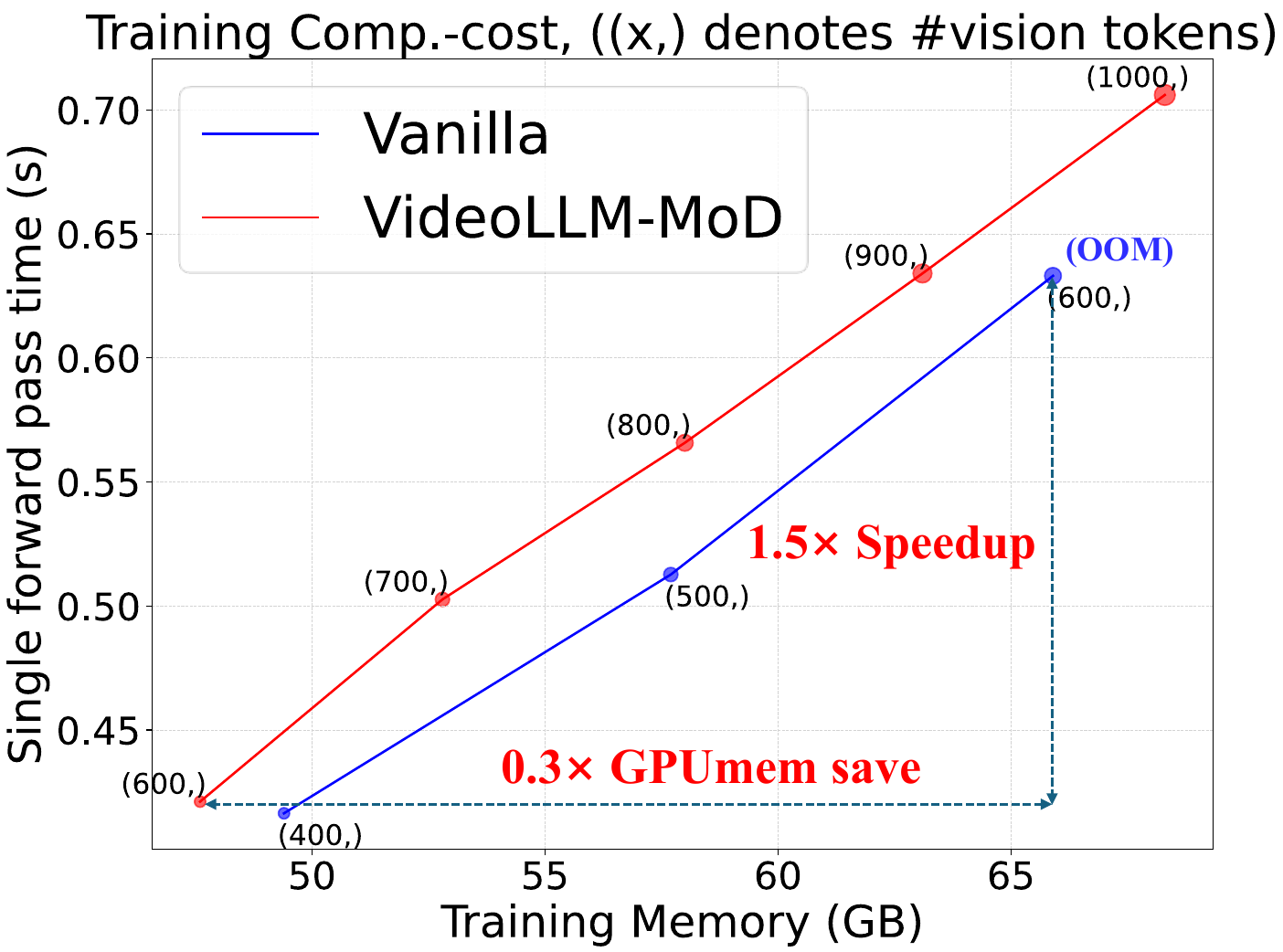} 
    \caption{Training Computation Cost. $\mymodel$ exhibits greater efficiency compared to the baseline.}
    \vspace{-5mm} 
    \label{fig:train_time_mem}
\end{wrapfigure}

It is intuitive to enhance spatial understanding by integrating additional pooled spatial tokens per frame,
as shown in Figure~\ref{fig:teaser}.
However, expanding vision resolution in the online scenario is challenging.
Due to the dense attention mechanism and the deep layer design of existing LLMs, the training cost, including GPU memory and training time, increases quadratically as the number of vision tokens expands (\textit{e.g.,} from 0.6k$\rightarrow$6k vision tokens for a video consisting of 600 frames), which poses significant challenges to scaling up vision capabilities.
Long videos, particularly online streaming videos, exhibit high redundancy in visual content, which suggests that a sparser approach could be used to process visual signals, potentially reducing the need for full attention in vanilla transformer-based LLMs without sacrificing performance.

In this paper, we propose $\mymodel$, an efficient approach to scaling up vision resolution for online video large language models.
Inspired by the Mixture-of-Experts (MoE)~\cite{moe}, which utilizes conditional logic to route tokens to one of many expert feed-forward networks (FFNs), we propose that vision tokens in specific blocks can be either routed to subsequent self-attention and FFN operations or bypassed via residual connections. This approach, which we term \textit{Mixture-of-Depth for vision tokens}, allows for the natural reduction of redundant vision information. Consequently, the model can learn which vision tokens are crucial, thereby optimizing computation accordingly. 
We surprisingly discovered that sparse operation at the token-level won't harm both vision capability and language modeling and is even better since it preserves original context as well as neglects the redundant vision signals, and it can dramatically reduce the training cost as shown in Figure~\ref{fig:train_time_mem}, serving as ``free lunch'' in vision scaling.
For each frame, instead of distributing FLOPs uniformly across all vision tokens in every decoder layer, we utilize a learnable module \texttt{LayerExpert} to allocate compute to critical vision tokens within the frame dynamically. Only a few vision tokens selected by the top-$k$ route mechanism are processed by the following self-attention and FFN (Feed Forward Network) operations and the remains are skipped through residual connection. 
Compared to directly dropping\cite{fastv,llava-prumerge} vision tokens or merging~\cite{blip2,chatunivi} them to reduce computation, the \textit{skip within context} mechanism preserves the completeness of the context, allowing for equivalent vision capability with significantly less computational effort.
The proposed approach can also be generalized to traditional offline video settings seamlessly, such as COIN~\cite{coin} and EgoExo4D~\cite{egoexo4d} benchmarks.

We summarize our technical contributions as follows:
\begin{itemize}[leftmargin=*,label=\textbullet]
    \item[$\bullet$] We propose $\mymodel$, an efficient approach to scaling vision resolution for online VideoLLM with reduced computational cost and similar or even better performance.
    \item[$\bullet$] We propose \texttt{LayerExpert} to determine which vision tokens should be processed at certain layers, leveraging the model to adaptively allocate computation to critical regions within incoming frames. 
    \item[$\bullet$] Our experiments on Ego4D, EgoExo4D, and COIN benchmarks demonstrate the effectiveness and generalizability of our $\mymodel$.
\end{itemize}
\section{Related Work}
\textbf{Efficient Modeling in Transformer-based Models.}
The notorious squared complexity in vanilla Transformers~\cite{transformer} is a well-known problem, as it is one of the key bottlenecks in scaling the sequence length. 
In recent large multimodal models (LMMs)~\cite{flamingo,blip2,llava,minigpt4,instructgpt}, prefix visual tokens are used as a fixed budget for context, significantly contributing to their efficiency. This issue becomes more pronounced in online video scenarios with denser video frames. 
Previous studies on large language models (LLMs)\cite{streamingllm,fastgen,fastv,layerskip} have explored the use of sparse computation to maintain performance during inference while reducing computational costs. However, these methods still incur significant training expenses. Efforts to reduce training costs through token pruning\cite{llava-prumerge} and merging~\cite{chatunivi} techniques are not suitable for online scenarios, as they require offline computation to directly reduce token levels.
Mixture-of-Depth~\cite{mod} investigates the allocation of computation across model depth for language tokens, balancing performance with speed. Our model, $\mymodel$, extends this approach to online video. We found that reducing vision computation in the context across model depth not only maintains but can even improve performance by removing high redundancy in video.

\textbf{Large Multimodal Models for Online Video Understanding.}
Inspired by the success of numerous large language models (LLMs)~\cite{gpt3,instructgpt,chatgpt}, a series of large multimodal models (LMMs)~\cite{flamingo,blip2,llava,minigpt4,instructblip} have subsequently been developed to further enhance our comprehension of the world.
Current large multimodal models (LMMs) are capable of addressing a variety of standard benchmarks in video understanding, including temporal action localization~\cite{stllm}, and video dialogue and question answering~\cite{videochat,moviechat,video_chatgpt,videollama,videollava}.
However, while these models analyze entire video frames to make predictions in an ``offline" setting, they are not optimized for real-time applications such as augmented reality (AR) glasses and autonomous driving systems. 
In light of this growing demand, benchmarks for online scenario such as action detection \cite{testra,oadtr} and anticipation \cite{avt,antgpt}, which are designed to interpret events at the current timestamp without access to future data, are becoming increasingly critical. 
VideoLLM-online~\cite{live} serves as the first attempt to build an assistant using LLMs in an online video scenario. However, its spatial capabilities are limited, as it uses only a single CLS token to represent each frame, and expanding the vision scale is computationally expensive. 
$\mymodel$ proposes an efficient approach to scaling vision resolution by reducing vision computation in context, thereby enhancing spatial ability without incurring high computational costs.

\textbf{Scaling up Vision Resolution for Large Multi-modal Models.}
Scaling up the visual resolution for LMMs is an effective approach to enhancing vision capabilities. By utilizing $5\times$ more vision tokens compared to LLaVA-1.5~\cite{llava1.5}, LLaVA-NeXT~\cite{llavanext} achieved improved vision understanding.
However, scaling vision tokens in online video scenarios presents significant challenges, as the training cost increases quadratically with the expansion of vision tokens, requiring the processing of every incoming frame in long videos. To handle long-context vision tokens in LMMs, CogAgent~\cite{cogagent} integrates high-resolution image features into a low-resolution pathway via cross-attention across decoder layers. LLaMA-VID~\cite{llamavid} utilizes context-attention to represent each frame with two key tokens. Both approaches are only applicable for offline video, as the high latency induced by the additional cross-attention mechanism is unacceptable in online scenarios.
In contrast, $\mymodel$ receives streaming video-language input continuously and can reduce computation efficiently during every forward pass without additional overhead. This enables temporal-aligned responses, making it suitable for real-time applications.

\section{Method}
In this section, we introduce our $\mymodel$ framework, an efficient approach to training an online video large language model with a larger vision resolution.

\subsection{Model architecture.}
We depict the overall model architecture as shown in Figure~\ref{fig:framework}, drawing parallels to LLaVA~\cite{llava,llava1.5,llavanext} in its design. The model is composed of three principal components: an image encoder, an MLP projector, and a language model. Each video frame embedding is represented as $(1+h_p \times w_p) \times c$, which denotes the CLS token and the average pooled spatial tokens.
The frame embeddings extracted by the image encoder are subsequently processed through the MLP projector to frame tokens. These tokens are interwoven with language tokens, forming the input for an LLM. We incorporate LoRA~\cite{lora} in every linear layer of the LLM for efficient tuning. 
To select the most critical vision tokens, certain layers are also equipped with \texttt{LayerExpert} module, as detailed in Section~\ref{sec:scale}.

Following VideoLLM-online~\cite{live}, in addition to the language modeling (LM) loss, we also utilize an additional streaming loss to ensure the model remains silent when it is unnecessary to output responses. Both training objectives employ cross-entropy loss as follows:

\begin{equation}
 L = \frac{1}{N}\sum^N_{j=1}(\underbrace{-l_{j+1}\log P_j^\texttt{[Txt$_{j+1}$]}}_{LM Loss} - \underbrace{\sigma s_j\log P_j^\texttt{[EOS]}}_{Streaming Loss}),
\end{equation}

where $l_j$ and $s_j$ are condition indicators: $l_j$ is 1 if the $j$-th token is a language response token, and 0 otherwise; $s_j$ is 1 if both (1) the $j$-th token is the \textit{last} token of a frame\footnote{Loss is applied only to the last token when a frame consists of multiple patch tokens.}, and (2) $l_{j+1} = 0$. The streaming EOS loss is applied to frames prior to responding. $P_j^\texttt{[Txt$_{j+1}$]}$ represents the probability associated with the $(j+1)$-th text token, as output by the language model head for the $j$-th token, while $P_j^{\texttt{[EOS]}}$ indicates that probability for the EOS token. The two objectives are balanced using the streaming loss weight $\sigma$.

\subsection{Scale up Vision Resolution for Online Video.} \label{sec:scale}
\textbf{Motivation.} The performance of online assistants improves with increased vision scale (\textit{i.e.,} $(1+h_p \times w_p)$), as shown in Figure~\ref{fig:teaser}. However, enhancing vision resolution in online scenarios is challenging because the number of vision tokens grows with video duration, leading to quadratic computational complexity. As shown in Table~\ref{tab:vision_strategy}, online videoLLMs must process every frame of long videos during both training and inference to maintain the integrity of the complete visual and linguistic historical contexts, which places significant demands on GPU memory and computational resources.

We hypothesize that videos exhibit high redundancy in temporal and spatial since consecutive frames often share a large portion of their content, especially if the frames are captured in quick succession. 
Just as humans continuously ``see" their surroundings without always ``focusing" on every visual detail, it is intuitive that we should skip some vision tokens in certain layers when processing them with a deep model.

\begin{figure}[h]
  \includegraphics[width=1.0\textwidth, height=0.55\textwidth]{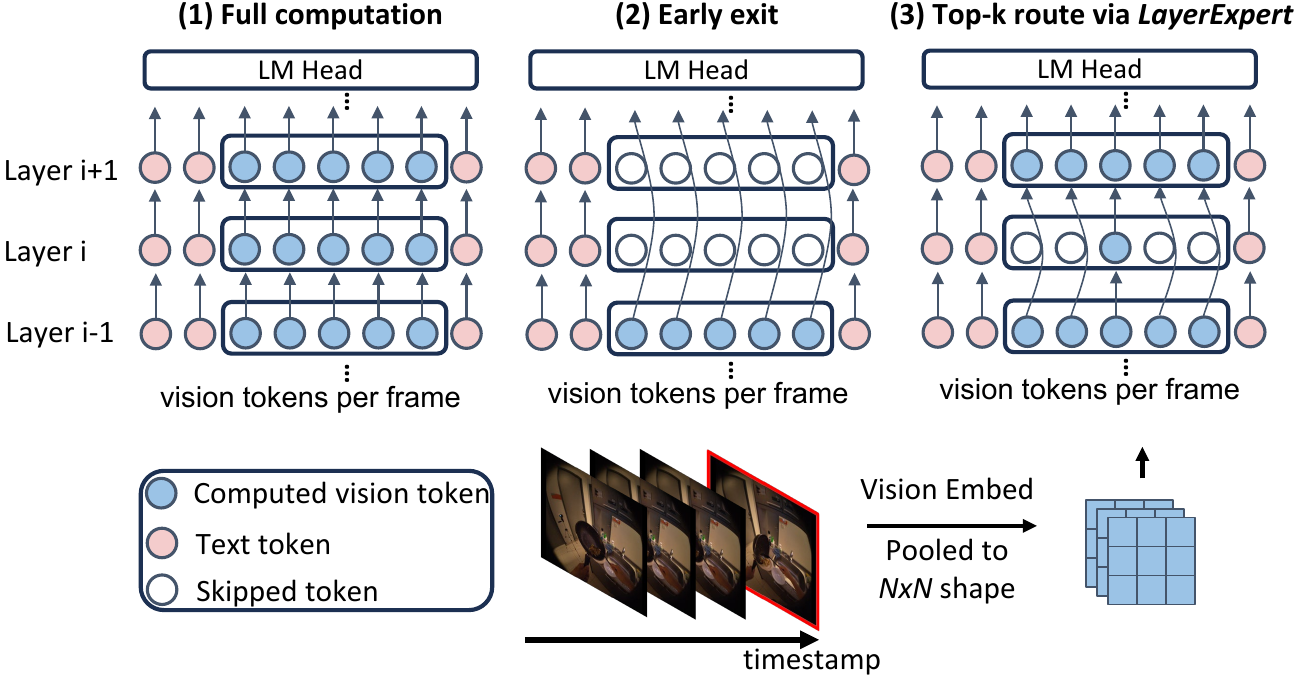}
  \caption{$\mymodel$ selects the top-$k$ vision tokens within each frame in certain layers via \texttt{LayerExpert}. We observe that performance drops dramatically with \textit{Early-exit} as critical vision tokens miss subsequent processing. By retaining crucial vision tokens in certain layers and reducing redundant tokens that may mislead understanding, $\mymodel$ achieves better performance with significantly lower computation costs compared to \textit{Full-computation} baseline.}
  \label{fig:illustration}
  \vspace{-3mm}
\end{figure}

\textbf{Selecting vision tokens via \texttt{LayerExpert} in certain block.}
We leverage a per-block \texttt{LayerExpert} module to learn the selecting/routing behavior, \textit{i.e.,} learn which vision tokens require more or less processing than others. The \texttt{LayerExpert} identify the ``importance score'' (in scalar weights) of each vision token within a frame, and only the top-$k$ vision tokens are processed by the following operations. 
Notably, since the vision tokens of different frames in the online streaming scenario are processed in a causal manner, the top-$k$ selection is performed at the frame level, meaning the top-$k$ vision tokens are selected within each frame.
The language tokens are always processed since they are much less redundant and significantly fewer in number compared to vision tokens.

Specifically, suppose we have a sequence of length $N$ interleaved with $n_t$ language tokens and $n_v$ vision tokens.
For the given layer $l$, the sequence $X^l = \{\text{Interleaved}(x_{t_i}^l, x_{v_i}^l) \mid 1 \leq t_i \leq n_t, 1 \leq v_i \leq n_v \}$.
Within the $(1+h_p \times w_p)$ vision tokens of each frame, the \texttt{LayerExpert} determines the importance score $\mu$ for a given vision token using a linear projection $\mu_{t_i}^l = w_\theta^T x_{v_i}^l$.
Then, vision tokens are selected based on a vision keep ratio $r$ for following processing, and $P_r^l$ is the (1-r)-th percentile among the weights $\mu$ of frame vision tokens. The block's output for the given vision token is as follows:

\begin{eqnarray}
    x_{v_i}^{l+1} = 
    \begin{cases} 
    \mu_{v_i} f_i(\mathbf{\hat{X}}^l) + x_{v_i}^l, & \text{if } \mu_{v_i} > P_r^l \\
    x_{v_i}^l, & \text{if } \mu_{v_i} \leq P_r^l
    \end{cases}    
\end{eqnarray}

where the $\hat{X}^l$ represents the interleaved tokens consisting of all language tokens and top-$k$ vision tokens in layer $l$, and $f_i$ denotes the subsequent self-attention and the FFN operations.

\subsection{Efficiency analysis of \textsc{$\mymodel$}.}

\begin{figure}[ht]
    \vspace{-8mm}
    \centering
    \begin{subfigure}[b]{0.45\textwidth}
        \centering
        \includegraphics[width=\textwidth]{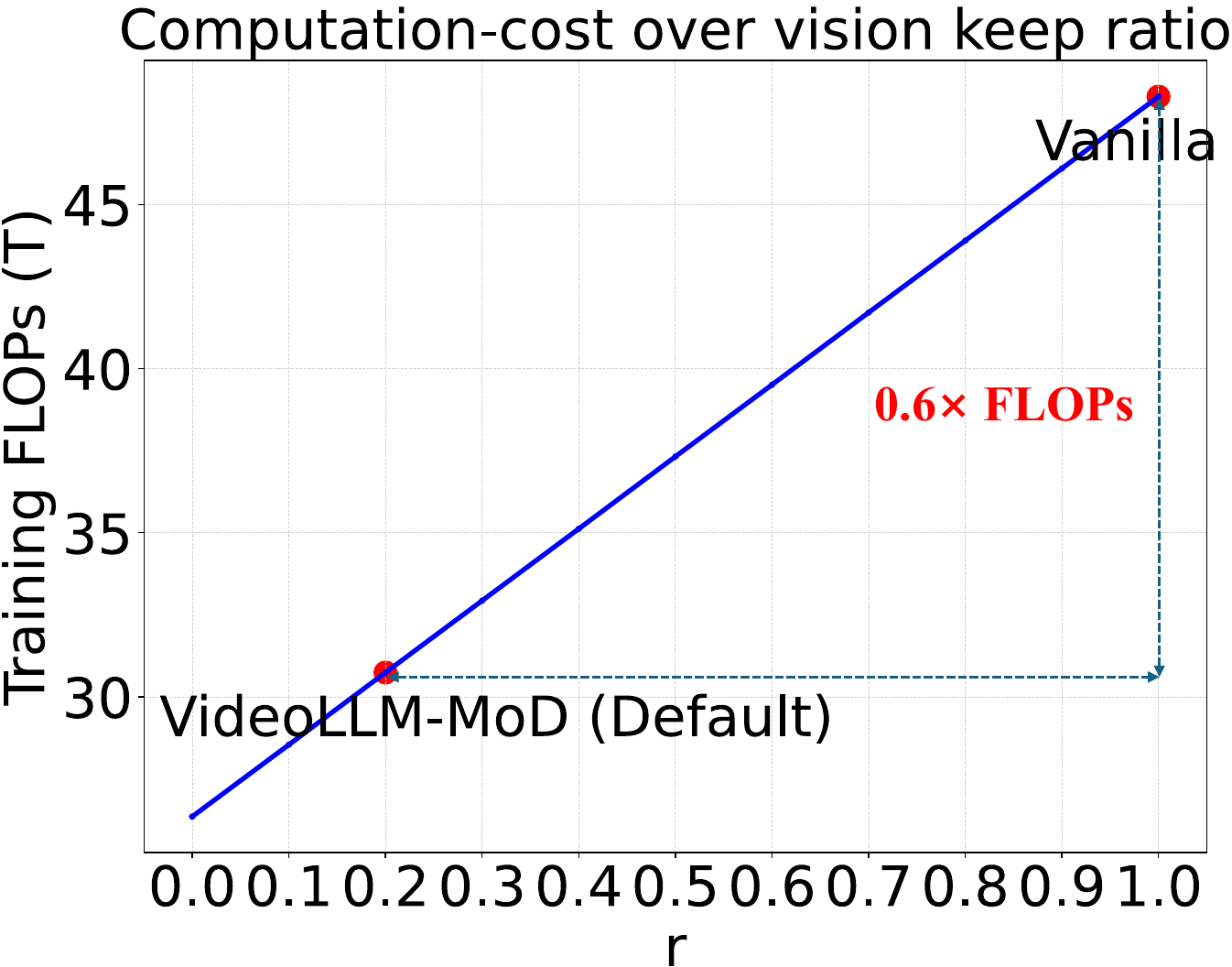}
        \caption{Calculated training FLOPs over vision keep ratio $r$. By default, $\mymodel$ only requires 0.6$\times$ FLOPs compared to the \textit{Full-computation} baseline.}
        \label{fig:train_flops}
    \end{subfigure}
    \hfill
    \begin{subfigure}[b]{0.45\textwidth}
        \centering
        \includegraphics[width=\textwidth]{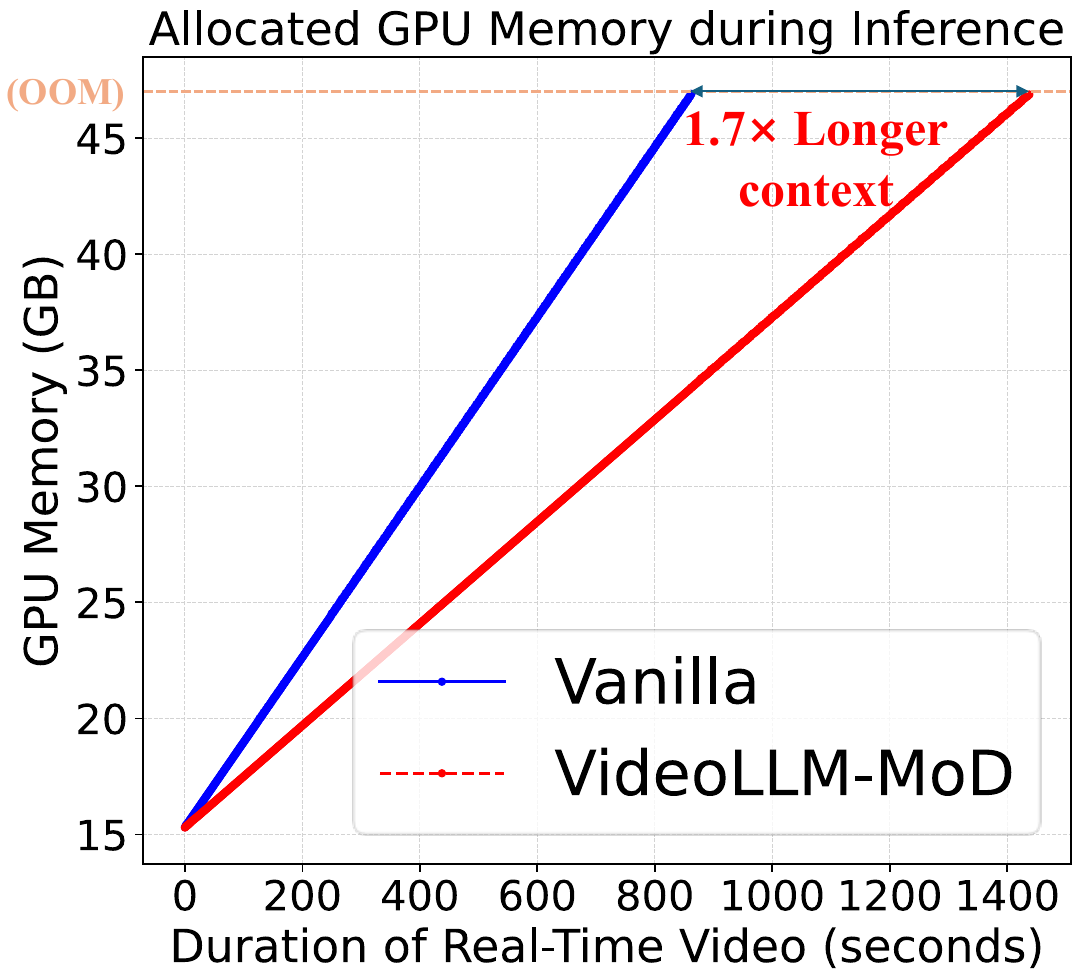} 
        \caption{Allocated GPUMemory in inference phase. By saving the KV cache of historical states, $\mymodel$ supports 1.7$\times$ longer video than baseline.}
        \label{fig:infer_mem}
    \end{subfigure}
    \caption{Efficiency analysis of $\mymodel$ in both training and inference phase.}
    \label{fig:efficiency_train_infer}
    \vspace{-5mm}
\end{figure}

We further analyze the computation cost of our approach.
Except for the decoder layers, the other modules, including \texttt{LayerExpert}, the MLP projector, and LoRA, are fixed given certain inputs and are significantly smaller than the decoder layers of the language model. Therefore, we ignore their FLOPs computation and only consider the computation of the multi-head attention (MHA) and feed-forward network (FFN) modules in the FLOPs estimation.

Suppose the language model has $L$ total hidden layers, in which $d$ and $m$ denote the hidden size dimension, and the intermediate size of FFN, respectively. 
The input sequence is interleaved with $n_v$ vision tokens and $n_t$ language tokens.
We insert \texttt{LayerExpert} in $K$ layers with a vision keep ratio $r$ inside of the entire decoder layers.
For each attention head, the theoretical FLOPs of the layer with \texttt{LayerExpert} is:

\begin{equation}
    \text{FLOPs}_{\text{LayerExpert}} = 4(n_t + r n_v)d^2 + 2(n_t + r n_v)^2 d + 2(n_t + r n_v)dm,
\end{equation}
while $r=1$ in vanilla transformer layers. Since in the online video setting, vision tokens are significantly more than language tokens, \textit{i.e.,} $n_v \gg n_t$, the FLOPs of entire decoder is proportional to the vision keep ratio $r$ and the number of layers equipped with \texttt{LayerExpert} as follows:

\begin{eqnarray}
    \text{FLOPs}_{\text{decoder}} = \sum_{k=K+1}^{L} \text{FLOPs}_{\text{LayerExpert}}(r) + \sum_{k=K+1}^{L} \text{FLOPs}_{\text{vanilla}}
\end{eqnarray}

We further calculate the total FLOPs\footnote{Calculated via DeepSpeed FLOPs Profiler, processing 600 frames with (1+3$\times$3) patches on a single NVIDIA A100 GPU. The layers with \texttt{LayerExpert} are interleaved in every other layer in default.} of $\mymodel$  and the \textit{Full-computation} baseline during the training phase in a real-world scenario. As shown in Figure~\ref{fig:train_flops}, the practical FLOPs of $\mymodel$ is only 0.6$\times$ that of the baseline, and this value can be further reduced if the vision scale of each frame is larger, demonstrating the excellent scalability of our approach.

By skipping redundant vision tokens in certain layers, $\mymodel$ not only reduces training computation costs but also improves inference efficiency. As shown in Figure~\ref{fig:infer_mem}, reducing the intermediate KVcache in historical states allows us to support 1.7$\times$ longer context and achieve a comparative inference speed compared to baseline, facilitating deployment in real-world applications.

\section{Experiments}
\subsection{Experimental Settings}
\textbf{Datasets.}
We validate the effectiveness of our proposed $\mymodel$ on both online and offline settings, including egocentric video dataset Ego4D~\cite{ego4d} and EgoExo4D~\cite{egoexo4d}, as well as instructional video dataset COIN~\cite{coin}.
\begin{enumerate}[leftmargin=*,label=\textbullet]
    \item \textbf{Ego4D Narration Stream Benchmark}: Following VideoLLM-online~\cite{live}, we utilize the dense Ego4D timestamp-narrations to create a streaming set, aiming to generate timely narrations similar to those produced by Ego4D human annotators~\cite{ego4d}. 
    \item \textbf{Ego4D long-term action anticipation (LTA) Benchmark}: This benchmark requires predicting the next $Z = 20$ actions (verbs and nouns) for a given video based on the previous 8 steps. Following previous studies~\cite{antgpt,palm}, we use the standard Ego4D v2 splits,     
    \item \textbf{EgoExo4D Fine-grained Keystep Recognition Benchmark}: This task involves recognizing fine-grained key steps from procedural egocentric videos during the test phase, using models that can leverage multiple time-synchronized views during training.
    \item \textbf{COIN Benchmarks}: Following previous works~\cite{distantsup,procedurevrl,videotf,unloc}, we evaluate our model on six common benchmarks of the COIN dataset: step recognition, step forecasting, task summarization, procedure forecasting, and procedure forecasting with a goal.
\end{enumerate}

\textbf{Evaluation metrics and implementation details}: For online benchmark, following VideoLLM-online~\cite{live}, we use the Language Modeling Perplexity (\textit{LM-PPL}) and \textit{LM-Correctness} to evaluate the language modeling capability at the given timestamp. To evaluate the temporal alignment capability as an online assistant, we use the Time Difference (\textit{TimeDiff}) and \textit{Fluency} to comprehensively evaluate both the language modeling and temporal effectiveness. We trained all models on 8$\times$ NVIDIA A100 GPUs. For each module, we use SigLIP-L/16~\cite{siglip} as the visual encoder, a 2-layer MLP as the multimodal projector, and Meta-Llama-3-8B-Instruct~\cite{llama3} for the language model. For the vision embedding of each video frame, we use $(1 + 3 \times 3)$ tokens (CLS token + averaged pooled spatial tokens), with a frame rate of 2 FPS. We add LoRA~\cite{lora} to all linear layers of the language model with a rank of 128 and a scaling factor of 256. Additional details can be found in Appendix~\ref{supp:details}. For the trade-off between computation cost and performance, we insert \texttt{LayerExpert} every other layer and set the keep ratio $r$ to 0.2 as the default setting.

\subsection{Online Experiments}
\begin{table*}[h]
\vspace{-3mm}
\setlength{\tabcolsep}{3pt}
\small
\centering
\begin{tabular}{l|c|c|c|ccccc}
\toprule
\multirow{2}{*}{Method} & \multirow{2}{*}{\makecell{Frame \\Strategy}} & \multirow{2}{*}{FLOPs} & Training Cost & \multicolumn{4}{c}{Ego4D Narration Stream on Validation Set} \\
& & & \& Speedup & \textit{LM-PPL}$\downarrow$  & \textit{TimeDiff}$\downarrow$ &  \textit{Fluency}$\uparrow$ & \textit{LM-Correctness}$\uparrow$ \\
\midrule
VideoLLM-online~\cite{live} & 1 & 5.75T & 8hrs \& n/a & 2.43 & 2.04 & 45.1\% & 48.1\% \\
\midrule 
Full-computation & \multirow{4}{*}{1+3$\times$3} & 48.29T & 24hrs \& n/a & 2.40 & 2.05 & 45.3\% & 49.0\% \\
EarlyExit && 7.14T & 10hrs \& 2.4$\times$ & 2.50 & 2.29 & 41.3\% & 46.2\% \\
LayerSkip && 26.35T & 13hrs \& 1.8$\times$ & 2.52 & 2.24 & 42.0\% & 46.5\% \\
\baseline{$\mymodel$} && \baseline{30.74T} & \baseline{14hrs \& 1.7$\times$} & \baseline{2.41} & \baseline{2.04} & \baseline{45.2\%} & \baseline{48.9\%} \\
\bottomrule
\end{tabular}
\caption{Online experiments on the Ego4D narration benchmark. $\mymodel$ achieves comparable metrics to the \textit{Full-computation} baseline with less computation cost.}
\vspace{-3mm}
\label{tab:ego4d_narration}
\end{table*}
We compare our \textsc{$\mymodel$} model with various baselines on the Ego4D narration benchmark, as shown in Table~\ref{tab:ego4d_narration}. We analyze the baselines in detail as follows.

\begin{figure}[h]
    \vspace{-3mm}
    \centering
    \includegraphics[width=14cm]{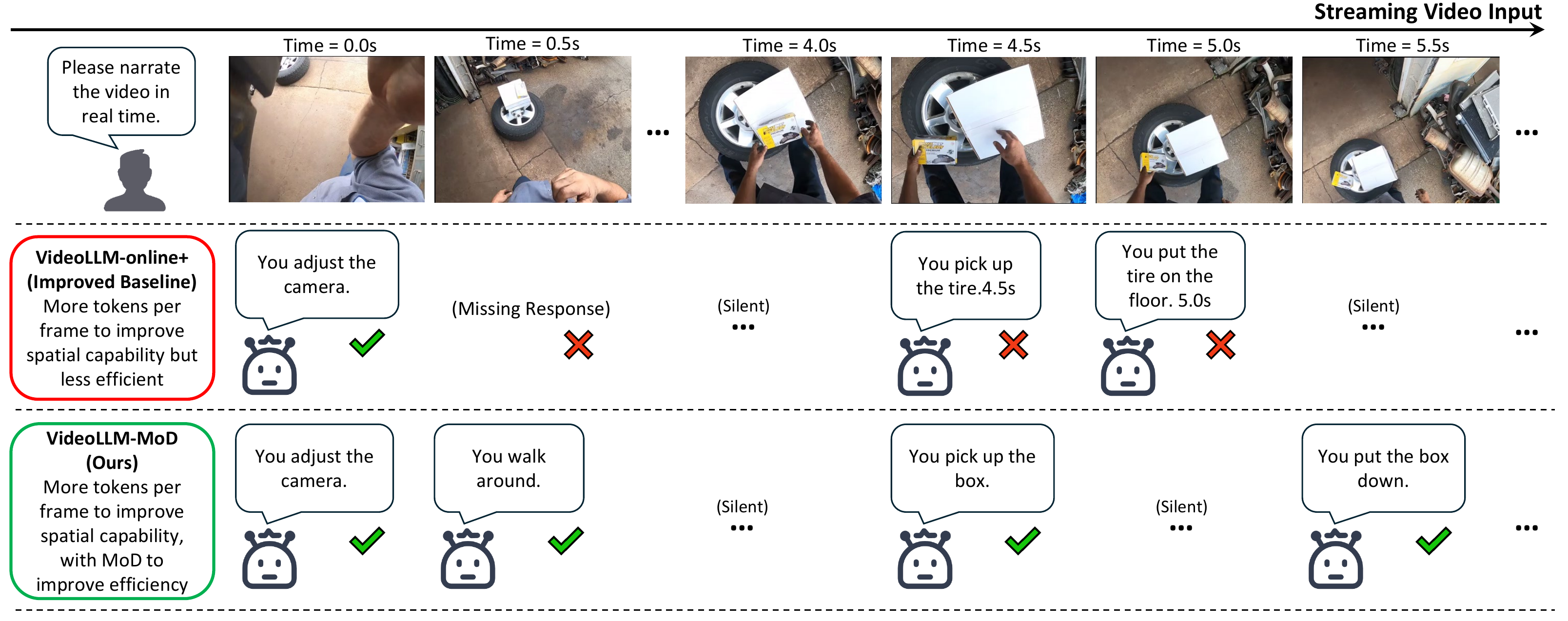}
    \caption{Examples of $\mymodel$ on the Ego4D GoalStep~\cite{ego4d_goalstep} video dataset. We found that $\mymodel$ effectively reduces hallucinations and performs more robustly than the model trained with full computation. For instance, our model correctly recognizes ``pick up the box" while the baseline mistakenly identifies it as ``pick up tire." Text in \textcolor{red}{red} indicates incorrect responses.}
    \label{fig:case}
    \vspace{-5mm}
\end{figure}

\textbf{VideoLLm-online.} For a fair comparison, we re-implemented the VideoLLm-online~\cite{live} baseline using the same visual encoder and language model as \textsc{$\mymodel$} in all experiments. By embedding each frame using only the CLS token, VideoLLm-online~\cite{live} achieves slightly worse performance on this benchmark due to the relatively simple narration, which does not heavily rely on fine-grained vision. Moreover, we found that larger vision resolution can indeed benefit performance, as shown in Figure~\ref{fig:teaser}, \ref{fig:case}, and in experiments that demand more detailed visual information as shown in Table~\ref{tab:coin_exp}, \ref{tab:offline_exp}.

\textbf{Full-computation.} Using a vanilla transformer architecture, all vision tokens are processed densely across every layer, which significantly increases the training cost.

\textbf{EarlyExit.} Building on studies of language-only LLMs~\cite{cvs,skipdecode}, we adapt this approach to the online video setting. All vision tokens are processed in the shallow layers, then skipped in the deeper layers (equivalent to \textsc{$\mymodel$} with $r=1$ in the first few layers and $r=0$ in the remaining layers). Empirically, we found that early exit at Layer 2 offers the best tradeoff between performance and computational cost, also highlighted in previous studies~\cite{streamingllm, fastv}. This approach shows the lowest computation but the worst performance, as it misses most of the vision information.

\textbf{LayerSkip.} Introduced in previous LLM studies~\cite{layerskip}, we adapted the approach to the online scenario, skipping all vision tokens in every other layer (treated as \textsc{$\mymodel$} interleaving layers with $r=1$ and $r=0$). Compared with \textsc{$\mymodel$}, the performance drops significantly as critical vision tokens miss processing in certain layers.

Our \textbf{\textsc{$\mymodel$}} exhibits the best tradeoff in online video scenarios, significantly reducing computational costs when processing excessive frames without sacrificing performance compared to the \textit{Full-computation} baseline. Moreover, we discovered that our approach performs better than the \textit{Full-computation} baseline in practical use, as shown in Figure~\ref{fig:case}.
It seems counterintuitive that fine-tuning LLM with MoD performs better than using the vanilla model. The \textit{dynamic layer skipping} methodology of the former results in less vision computation during the forward process, which is likely to weaken the spatial understanding capability. However, we argue that this increases the learning difficulty, as it forces the MoD gate at each layer to focus on the important vision tokens in current causal contexts. This may reduce the risk of overfitting and learn a more robust model.

\subsection{Ablation Study}
\textbf{Insertion strategy of \texttt{LayerExpert}.}

Table~\ref{tab:ablation_layerexpert} presents the ablation study results for the insertion strategies of our \texttt{LayerExpert}.
\begin{wraptable}{r}{0.45\textwidth}
\vspace{-3mm}
  \begin{minipage}{0.47\textwidth}
    \centering 
    \scriptsize
    \setlength{\tabcolsep}{1pt}
      \centering
        \resizebox{!}{1.1cm}{
        \begin{tabular}{cccc}
            \toprule
            Insertion strategy & \textit{Fluency}$\uparrow$& \textit{LM-Correctness}$\uparrow$ & FLOPs \\
            \midrule
             All & 42.5\% & 47.5\% & 13.18T \\
             All-Deep & 44.8\% & 48.8\% & 15.37T \\
             \baseline{Interleaved} & \baseline{45.2\%} & \baseline{48.9\%} & \baseline{30.74T} \\
             Interleaved-Deep & 45.2\% & 49.0\% & 31.83T \\
            \bottomrule
        \end{tabular}
        }
    \caption{Ablations on the insertion strategy of \texttt{LayerExpert} in transformer layers. The \textit{Interleaved} strategy strikes the best trade-off among the variants.}
    \label{tab:ablation_layerexpert}   
    \vspace{-5mm}
\end{minipage}
\end{wraptable}
We constructed different settings for inserting \texttt{LayerExpert} in the transformer layers. \textit{All} and \textit{Interleaved} refers to insert \texttt{LayerExpert} in every/every other layer.
The \textit{Interleaved} strategy demonstrates a better trade-off between computation cost and performance.

The postfix \textit{-Deep} denotes that vision token skipping is performed only in the deep layers (\textit{i.e.,} layers after Layer 2). Previous studies~\cite{streamingllm, fastv} indicate that attention allocation across all tokens in the shallow layers (the first two layers) is much more balanced compared to the deep layers, making these shallow layers more vulnerable to token skipping. Our results with and without \textit{-Deep} also indicate this phenomenon.

\textbf{Selecting the critical vision token.}
As shown in Table~\ref{tab:ablation_allocation}, to validate the necessity and effectiveness of allocating computation to the crucial vision tokens, we created two variants that select vision tokens either randomly or uniformly. 
The poorer performance on \textit{TimeDiff} indicates that the online capability is significantly impacted by missing critical vision information. This suggests that determining which vision tokens deserve processing is essential for maintaining performance while reducing redundancy. 
\begin{wraptable}{r}{0.65\textwidth}
\vspace{-2mm}
  \begin{minipage}{0.65\textwidth}
      \centering
        \resizebox{!}{1.3cm}{
            \begin{tabular}{l|l|cccc|c}
            \toprule
            \multirow{2}{*}{\makecell{Keep\\Strategy}} & \multirow{2}{*}{\makecell{Keep\\Ratio $r$}} & \multicolumn{4}{c}{Ego4D Narration Stream Validation} \\
            &&\textit{LM-PPL}$\downarrow$   & \textit{TimeDiff}$\downarrow$ &  \textit{Fluency}$\uparrow$ & \textit{LM-Correctness}$\uparrow$ & FLOPs\\
            \midrule
            Random & $r = 0.2$ & 2.45 & 2.18 & 43.6\% & 48.1\% & 30.74T \\ 
            \midrule
            Uniform & $r = 0.2$ & 2.42 & 2.17 & 43.9\% & 48.6\% & 30.74T \\ 
            \midrule
            \multirow{3}{*}{Learnable} & $r = 0.1$ & 2.43 & 2.11 & 44.7\% & 48.1\% & 28.54T \\
            & \baseline{$r = 0.2$} & \baseline{2.41} & \baseline{2.04} & \baseline{45.2\%} & \baseline{48.9\%} & \baseline{30.74T} \\
            & $r = 0.3$ & 2.41 & 2.05 & 44.9\% & 48.7\% & 32.93T \\
            \bottomrule
            \end{tabular}
            }
            \caption{Ablations on different vision selection strategies. Choosing which vision tokens to process is crucial for efficient vision computation allocation.}
            \label{tab:ablation_allocation}
    \end{minipage}
\vspace{-3mm}
\end{wraptable}
We also conducted ablations on the number of vision tokens to retain based on the vision keep ratio $r$. Even with relatively fewer tokens and FLOPs, \textsc{$\mymodel$} achieves satisfactory results, further demonstrating the critical vision selection capability of \texttt{LayerExpert} and highlighting the high redundancy present in the video.

\subsection{Offline Experiments}
\vspace{-3mm}
\begin{wraptable}{r}{0.65\textwidth}
\vspace{-5mm}
  \begin{minipage}{0.65\textwidth}
\small
\centering
\resizebox{!}{1.9cm}{
    \begin{tabular}{l|c|ccccc}
    \toprule
    \multirow{2}{*}{Method} & \multirow{2}{*}{\shortstack{Not use\\HowTo100M}} & \multicolumn{5}{c}{COIN Benchmark Top-1 Accuracy$\uparrow$} \\
    & & Step & Task & Next & Proc. & Proc.+ \\
    \midrule
    ClipBERT~\cite{clipbert}  & \checkmark & 30.8 & 65.4 & - & - & - \\
    TimeSformer~\cite{timesformer}  & \xmark & 46.5 & 85.3 & 34.0 & 17.0 & 40.1 \\
    Paprika~\cite{paprika}   & \xmark & 51.0 & 85.8 & 43.2 & - & - \\
    DistantSup~\cite{distantsup}   & \xmark & 54.1 & 90.0 & 39.4 & - & 41.3 \\
    VideoTF~\cite{videotf} & \xmark & 56.5 & 91.0 & 42.4 & 40.2 & 46.4 \\
    ProcedureVRL~\cite{procedurevrl}  & \xmark & 56.9 & 90.8 & 46.8 & - & - \\
    VideoTaskGraph~\cite{video_mined_task_graph} & \xmark & 57.2 & 90.5 & 40.2 & - & - \\
    VideoLLM-online~\cite{live} & \checkmark & 62.5 & 92.2 & 49.3 & 48.6 & 53.3 \\
    Ours (Full-computation) & \checkmark & 63.1 & 92.7 & 49.1 & 49.8 & \textbf{54.1} \\
    \baseline{Ours} & \baseline{\checkmark} & \baseline{\textbf{63.4}} & \baseline{\textbf{92.8}} & \baseline{\textbf{49.7}} & \baseline{\textbf{49.8}} & \baseline{53.3} \\
    \bottomrule
    \end{tabular}
}
\caption{Results on COIN benchmarks (left to right): step recognition, task recognition, next forecasting, procedure forecasting, procedure forecasting with a goal.}
\label{tab:coin_exp}
\end{minipage}
\end{wraptable}
We demonstrate the generalizability of our proposed \textsc{$\mymodel$} on traditional offline video scenarios, including recognition, summarization, and forecasting tasks. 
As shown in Table~\ref{tab:lta}, our method achieves the best performance compared to end-to-end baselines on the Ego4D LTA benchmark, with results only slightly lower than Palm~\cite{palm} and AntGPT~\cite{antgpt}, which utilize EgoVLP~\cite{egovlp} pretrained features followed by cascading performance enhancement methods.

By expanding the vision resolution, \textsc{$\mymodel$} achieves state-of-the-art performance, significantly surpassing videoLLM-online~\cite{live}, which only uses the CLS token for each frame. This is particularly evident in tasks requiring complex spatial context understanding, such as the EgoExo4D Fine-grained Keystep recognition benchmark~\cite{egoexo4d}, as shown in Table~\ref{tab:egoexo4d_keystep}.
Furthermore, our method achieves the best performance on most of the COIN benchmarks~\cite{coin}, as illustrated in Table~\ref{tab:coin_exp}, even outperforming our full-computation baseline.
By adaptively focusing on processing critical vision tokens, \textsc{$\mymodel$} not only reduces computation but also excels in spatial-temporal scene understanding, particularly in complex contexts.

\begin{table}[t]
    \vspace{-8mm}
    \begin{subtable}{.5\linewidth}
    \setlength{\tabcolsep}{2pt}
      \centering
        \resizebox{!}{2.1cm}{
            \begin{tabular}{l|l|l|ccc}
            \toprule
            \multirow{2}{*}{Method} & \multirow{2}{*}{\shortstack{Not use\\EgoVLP}} & \multirow{2}{*}{\shortstack{End-to\\-end?}} & \multicolumn{3}{c}{Ego4D LTA ED@Z=20$\downarrow$} \\
            & & & Verb & Noun & Action \\
            \midrule
            CLIP~\cite{video+clip4lta} & \checkmark & \checkmark &0.739&0.769&0.941\\
            EgoT2~\cite{egot2} & \checkmark &\checkmark&0.722&0.764&0.935 \\
            I-CVAE~\cite{icvae} &\checkmark & \checkmark &0.753&0.749&0.931\\
            HierVL~\cite{hiervl} &\checkmark & \checkmark & 0.724&0.735&0.928\\
            VideoLLM~\cite{videollm}  & \xmark & \checkmark & 0.721 & 0.725 & 0.921 \\
            Vi.LLM-o.~\cite{live} & \checkmark & \checkmark & 0.691 & 0.692 & 0.889 \\
            \baseline{Ours} & \baseline{\checkmark} & \baseline{\checkmark} & \baseline{\textbf{0.689}} & \baseline{\textbf{0.676}} & \baseline{\textbf{0.884}} \\
            \midrule
            \textcolor{gray}{Palm~\cite{palm}} & \xmark & \xmark  &\textcolor{gray}{0.696} &\textcolor{gray}{0.651} & \textcolor{gray}{0.886}\\
            \textcolor{gray}{AntGPT~\cite{antgpt}} & \xmark & \xmark &\textcolor{gray}{0.650}&\textcolor{gray}{0.650}&\textcolor{gray}{0.877}\\
            \bottomrule
            \end{tabular}
        }
    \caption{Results on Ego4D LTA benchmark, evaluated on \href{https://eval.ai/web/challenges/challenge-page/1598/evaluation}{public server}. ED@Z=20 denotes editing distance for future 20 actions.}
    \label{tab:lta}
    \end{subtable} 
    \begin{subtable}{.5\linewidth}
    \setlength{\tabcolsep}{1pt}
      \centering
        \resizebox{!}{2.4cm}{
            \begin{tabular}{l|l|c}
            \toprule
            Method & Train data (v1) & \makecell{Ego Accuracy (\%) \\on Val (v1/v2)} \\
            \midrule
            TimeSFormer~\cite{timesformer} (K600) & ego & 35.25/35.13 \\ 
            TimeSFormer~\cite{timesformer} (K600) & ego, exo & 32.67/32.68 \\ 
            EgoVLPv2~\cite{egovlpv2} (Ego4D) & ego & 36.89/36.51 \\
            EgoVLPv2~\cite{egovlpv2} (Ego4D) & ego, exo & 37.03/35.84  \\
            EgoVLPv2~\cite{egovlpv2} (EgoExo) & ego & 37.61/36.04  \\ 
            EgoVLPv2~\cite{egovlpv2} (EgoExo) & ego, exo & 38.21/39.10 \\ 
            VI Encoder~\cite{viencoder} (EgoExo)  & ego, exo & 40.23/40.34   \\ 
            Viewpoint Distillation~\cite{distillation}& ego, exo & 37.79/38.19   \\ 
            Ego-Exo Transfer MAE~\cite{egoexomae} & ego, exo & 36.71/37.17 \\ 
            VideoLLM-online~\cite{live} & ego & 40.53/40.73 \\
            \baseline{Ours} & \baseline{ego} & \baseline{\textbf{44.85}/\textbf{42.62}} \\
            \bottomrule
            \end{tabular}
        }
        \caption{Results on EgoExo4D Fine-grained Keystep Recognition benchmark.}
        \label{tab:egoexo4d_keystep}
    \end{subtable}%
\caption{Experiments on COIN~\cite{coin} and Ego4D~\cite{ego4d} benchmarks. VideoLLM-online is finetuned on their training set, and strictly evaluated on the test set by generated string comparison with the ground-truth text. It achieves best results among end-to-end models.
\vspace{-5mm}
}
\label{tab:offline_exp}
\end{table}

\subsection{Visualization of $\mymodel$.}
We observe that $\mymodel$ performs more robustly compared to the \textit{Full-computation} baseline. By selecting critical vision tokens, the learning difficulty increases, as this approach forces the MoD gate at each layer to focus on important vision tokens within current causal contexts. This strategy may reduce the risk of overfitting, thereby resulting in a more robust model.

\section{Conclusion, Limitations, and Broader Impacts}
In this paper, we introduced $\mymodel$, which scales vision resolution for video large language models in online video through efficient select critical video tokens via \texttt{LayerExpert}. Our model can significantly reduce computational costs and memory usage with similar or even better performance compared with \textit{Full-computation} baseline. Experiments on Ego4D, EgoExo4D, and COIN benchmarks confirm its efficacy and generalizability, making $\mymodel$ a robust solution for online video applications.


\textbf{Limitations.} As our primary focus was on developing an online assistant for egocentric or instructional scenarios, we did not conduct extensive experiments on exocentric video datasets. 

\textbf{Broader Impacts.} Beyond the online scenario, we hope our work can provide insights and contribute to general video understanding tasks, particularly those involving long videos.



\clearpage

\bibliography{ai}
\bibliographystyle{plain}  

\newpage
\appendix
\section{Appendix}
\subsection{Online Video Demo.}
We made several demo videos to showcase $\mymodel$'s effectiveness, available on the anonymous website \href{https://sites.google.com/view/videollm-mod-anonymous}{https://sites.google.com/view/videollm-mod-anonymous}.

\subsection{Model Architecture}
\begin{figure}[h]
  \includegraphics[width=1.0\textwidth, height=0.55\textwidth]{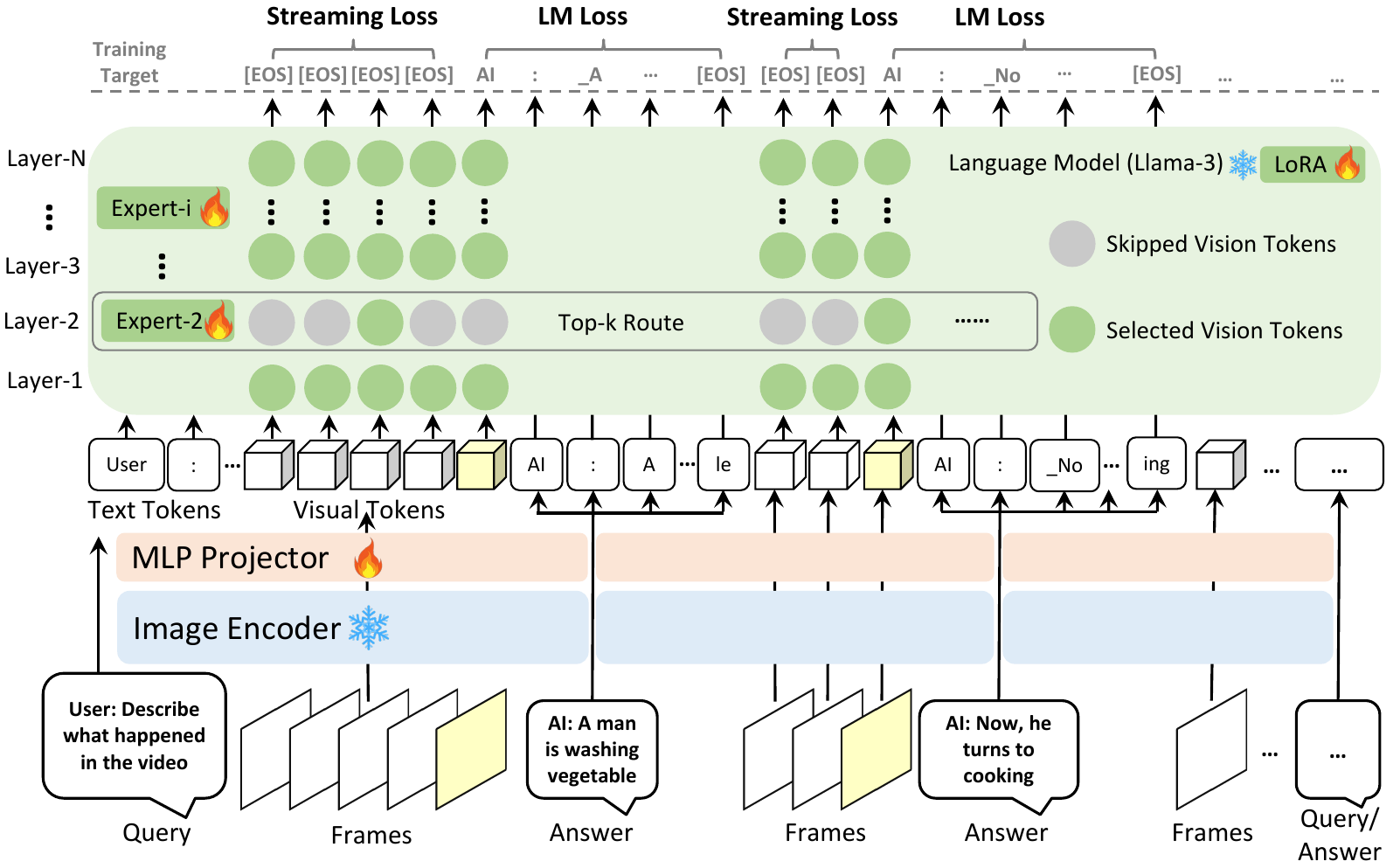}
  \caption{Model architecture of $\mymodel$, \img{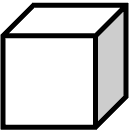} \img{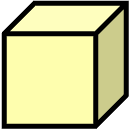} \img{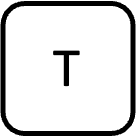} represent vision, keyframe, and text tokens, respectively.}
  \label{fig:framework}
\end{figure}

We depict the overall model architecture as shown in Figure~\ref{fig:framework}, which is composed of three principal components: an image encoder, an MLP projector, and a language model. Following VideoLLM-online, we train the model with both language model loss and streaming loss.

\subsection{Implementation Details} \label{supp:details}
For a fair comparison, we trained the models on the Ego4D narration benchmark for 2 epochs with a learning rate of $2 \times 10^{-4}$. For the Ego4D LTA benchmark, EgoExo4D fine-grained keystep recognition benchmark, and Coin benchmark, we trained the models for 6, 10, and 5 epochs with learning rates of $3 \times 10^{-4}$, $2 \times 10^{-4}$, and $1 \times 10^{-4}$, respectively. During training, we set the batch size to 64 and streaming loss weight $\sigma$ to 1.0 by default. 
For the trade-off between computation cost and performance, we insert \texttt{LayerExpert} every other layer and set the keep ratio $r$ to 0.2 as the default setting.

\subsection{Vision Strategy in Popular Large Multimodal Models.}
We summarize the vision strategy employed in popular large multimodal models (LMMs) as shown in Table~\ref{tab:vision_strategy}. The number of vision tokens increases continuously with video duration, as online video LMMs demand processing for every frame of the entire video. This leads to a quadratic increase in computational requirements during training and inference.
\begin{figure}[t]
    \centering 
    \scriptsize
    \setlength{\tabcolsep}{1pt}
\scalebox{1.3}{
    \begin{tabular}{c|l|c|c}
    \toprule
    \multicolumn{2}{c|}{Different LMMs} & TemporalSample & \#Token/frame$\rightarrow$\#Total VTokens \\
    \midrule
    \multirow{7}{*}{\makecell{Image-based\\(single image)}} & BLIP-2~\cite{blip2} & - & $32 \rightarrow 32$ \\
    & LLaVA~\cite{llava} & - & $256\rightarrow256$\\
    & LLaVA-1.5~\cite{llava1.5} & - & $576\rightarrow576$ \\
    & LLaVA-NEXT~\cite{llavanext} & - & $5\times576\rightarrow2880$ \\
    & DeepSeek-VL~\cite{deepseekvl} & - & $576\rightarrow576$ \\
    & Idefics2~\cite{idefics2} & - & $5\times64\rightarrow320$ \\
    & GPT-4V~\cite{gpt4} & - & $765$ (for $1024^2$ Res.)\\
    \midrule
    \multirow{4}{*}{\makecell{Offline video-based\\$(5\sim35s)$}} & Video-LLaVA~\cite{videollava} & 8 & $256\rightarrow 2048$ \\
    & VideoChat2~\cite{videochat2} & 8 & $96\rightarrow 768$ \\
    & Chat-UniVi~\cite{chatunivi} & \multicolumn{2}{c}{merge to $112$ S-T tokens} \\
    & Video-ChatGPT~\cite{video_chatgpt} & \multicolumn{2}{c}{merge to $356$ S-T tokens} \\ 
    \midrule
    \multirow{3}{*}{\makecell{Online/long video-based\\($\sim300s$, e.g., 2fps)}} & LLaMA-VID~\cite{llamavid} & Use all & $2\rightarrow 1.2k$ (LongLoRA~\cite{longlora} tuned) \\
    & VideoLLM-online~\cite{live} & Use all & $1 \rightarrow 0.6k$\\
    & \baseline{$\mymodel$} & \baseline{Use all} & \baseline{$1+3\times3 \rightarrow 6k$} \\
    \bottomrule
    \end{tabular}
}
    \vspace{2mm}
    \caption{Vision strategy in popular LMMs.}
    \label{tab:vision_strategy}  
\end{figure}


\subsection{Online correction via data augmentation on temporal.}

During streaming inference, we find that the model may be overfitted to the previous text output, likely due to insufficient training data in the LiveChat~\cite{live} dataset. To address this problem, we introduce a simple data augmentation strategy that requires no additional data annotation but, in our observation, significantly alleviates overfitting. The core idea is to disrupt the language context during streaming video-language modeling, reducing the model's bias towards language alone. Our ``disruption'' method involves randomly shifting within a temporal window and randomly replacing text with text from other timestamps within the same video data, or both. For the disrupted text, we do not perform language modeling or streaming EOS modeling on its adjacent video frames. 

All figures presented in this paper were obtained using this augmentation method. We will further investigate whether this approach can mitigate the shortage of high-quality streaming video-language data.

\end{document}